**DEVELOPMENT OF A BIOMECHANICAL MOTION SENSORIMOTOR PLATFORM FOR ENHANCED LOCOMOTION UNDER MICROGRAVITY CONDITIONS.** P.A. Johnson[1,2], J.C. Johnson[1,2], L. Tombrowski[1], S. Zirnov[1], R. Witiw[1,2], A.A. Mardon[1,2]. [1]The Antarctic Institute of Canada (11919- 82 Street NW, Edmonton, Alberta, Canada, aamardon@yahoo.ca). [2]University of Alberta (116 St 85 Ave NW, Edmonton, Alberta, Canada, jcj2@ualberta.ca; paj1@ualberta.ca)

**Introduction:** For humans accustomed to 1-G environments on Earth, microgravity conditions in orbit and on celestial bodies with lower gravitational field, such as the Moon, can be physiologically compromising. Of these, motor and fine-dexterity tasks involving the extremities, particularly in locomotion, grasp and release, are influenced becoming delayed and placing greater force demands. With the accelerating pace of prosthesis developments, research has reached frontiers in the development of biomechanical systems providing both sensory and motor feedback platforms to the user. The authors hereby propose incorporating this same technological innovation into loading suits designed for use in orbit or celestial environment.

**Prosthetic biomechanical model**: Current prosthetic systems use electromyography (EMG)-based techniques for creating functional sensorimotor platforms. This model sustains a sensorimotor platform for prosthesis users (Fig. 1).

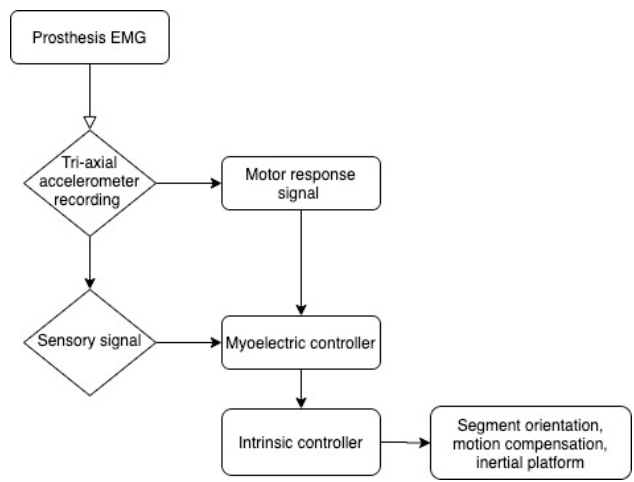

**Figure 1**. Tri-axial sensorimotor prosthesis model

However, several limitations in practical use and signal detection have been identified in these systems. Accelerometer-based sensorimotor systems have been suggested to overcome these limitations but only proof-of-concept has been demonstrated. Kyberd and Poulton have described a tri-axial system whereby sensors and controllers are employed to detect and correct for key motor control elements consisting of 1) segment orientation, 2) motion compensation, and 3) inertial platform [1]. Segment orientation is a compensatory mechanism for the accelerometer that takes into consideration the gravitational forces and tri-dimensional, spatial alignments in order to accommodate the motor demand accordingly. Alternatively, motion compensation adapts for the positioning using the surrounding prosthetic limb segment kinematics. In addition, the inertial platform controller uses holistic, mathematical analysis of prosthesis in interaction with an object of interest.

**Loading suit incorporation model**: Our group previously suggested the combined application of prosthetic accelerometers in loading suits and EMGs for input signal detection, quantification, and predictive output modeling necessary for improved motion adjustments under microgravity conditions [2]. Gravitation is not expected to have an effect, as this system exploits the Equivalence Principle, which states forces due to gravity and acceleration are indistinct. In other words, the gravitational acceleration or otherwise lack of gravity would not affect this feedback system. Additionally, as EMGs are the sole input in this model, complex multi-input variable modeling, which are quite common in biosensor feedback systems, can be avoided. It is anticipated this technology can enhance tasks such as repairs or construction or perhaps in recreational design when considering commercial and private human access to space.

*Limitations*. Though conceivable, feasibility and a proof-of-concept must be demonstrated prior to implementation. Foreseeable limitations with this design include establishing a differential feedback transduction system design for non-prosthetic users and ergonomic considerations for loading suits. Additionally, surface EMGs are limited by signal noise, when compared to needle electrode and fine-wire EMGs, which are often classified as gold standards for signal detection for diagnostic purposes. There additionally exists the need to design this within currently costly loading suit designs, which may not be economical in the context of resources and costs, especially for large-scale space flights or missions.

**Conclusion**: The incorporation of a prosthetic biomechanical sensorimotor system in loading suits to enhance locomotion under microgravity conditions are conceivable; however, demonstration of a proof-of-concept is required before implementation.

**References**: [1] Kyberd P. J. and Poulton A. (2017) IEEE Trans Neural Syst Rehabil Eng, 25(10), 1884– 1890. [2] Johnson P. A. *et al.* (2019) LPI Contribution No. 2152, id.5109.